\documentclass[runningheads]{llncs}
\usepackage{graphicx}
\usepackage{multirow}
\usepackage{caption}
\usepackage{cite}
\usepackage{mathtools}
\usepackage{subfigure}
\usepackage{array}
\usepackage{booktabs}
\usepackage{footnote}

\newcommand{\tabincell}[2]{\begin{tabular}{@{}#1@{}}#2\end{tabular}}
\begin{document}
\title{PEL-BERT: A Joint Model for Protocol Entity Linking}
%
%
\author{
Shoubin Li\inst{1,2} \and
Wenzao Cui \inst{3} \and
Yujiang Liu\inst{2}\and
Xuran Ming \inst{4} \and
Jun Hu\inst{2} \and
Yuanzhe Hu\inst{1,2} \and
Qing Wang\inst{2}
}
%
%
\institute{
University of Chinese Academy of Sciences, China \and
The Institute of Software, Chinese Academy of Sciences, China \and
Sichuan University, China \and
Beijing Jiaotong University, China\\
\email{\{shoubin,yujiang,hujun,yuanzhe,wq\}@iscas.ac.cn,wenzac77@outlook.com,Xuran.ming@bjtu.edu.cn}
}
\maketitle              

\begin{abstract}
Pre-trained models such as BERT are widely used in NLP tasks and are fine-tuned to improve the performance of various NLP tasks consistently. Nevertheless, the fine-tuned BERT model trained on our protocol corpus still has a weak performance on the Entity Linking (EL) task. In this paper, we propose a model that joints a fine-tuned language model with an RFC Domain Model. Firstly, we design a Protocol Knowledge Base as the guideline for protocol EL. Secondly, we propose a novel model, PEL-BERT, to link named entities in protocols to categories in Protocol Knowledge Base. Finally, we conduct a comprehensive study on the performance of pre-trained language models on descriptive texts and abstract concepts. Experimental results demonstrate that our model achieves state-of-the-art performance in EL on our annotated dataset, outperforming all the baselines.


\end{abstract}
%
%
\section{Introduction}
\begin{figure}[htb]
\setlength{\abovecaptionskip}{0pt}
\setlength{\belowcaptionskip}{10pt}
\includegraphics[width=\textwidth]{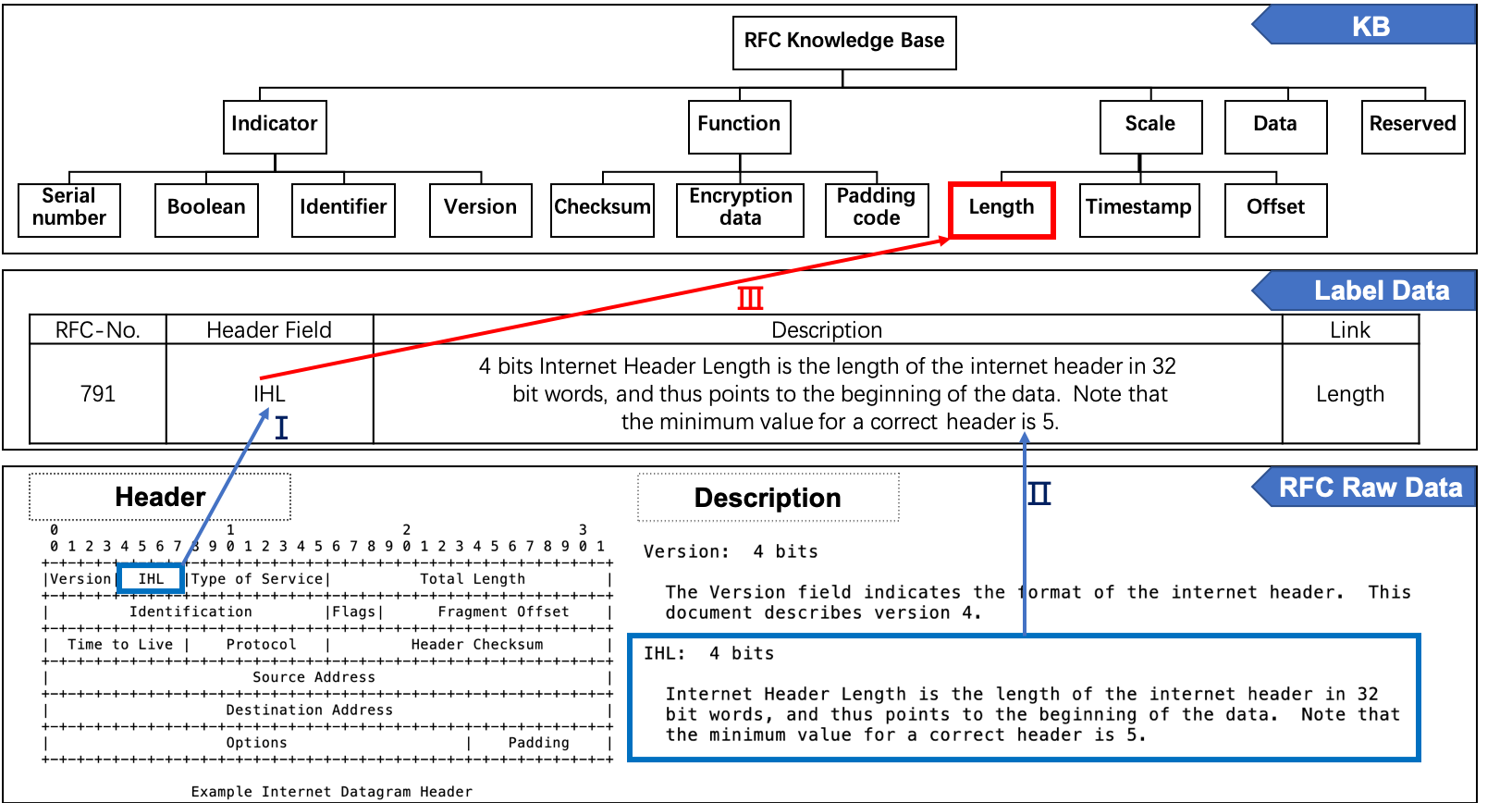}
\caption{Overview of Entity Linking in RFCs.
{\scriptsize \uppercase\expandafter{\romannumeral1}. Entity Extraction.
\uppercase\expandafter{\romannumeral2}. Context Inference.
\uppercase\expandafter{\romannumeral3}. Entity Linking.}
}\label{fig1}
\end{figure}

\begin{figure}[htb]
\setlength{\abovecaptionskip}{0pt}
\setlength{\belowcaptionskip}{20pt}
\includegraphics[width=\textwidth]{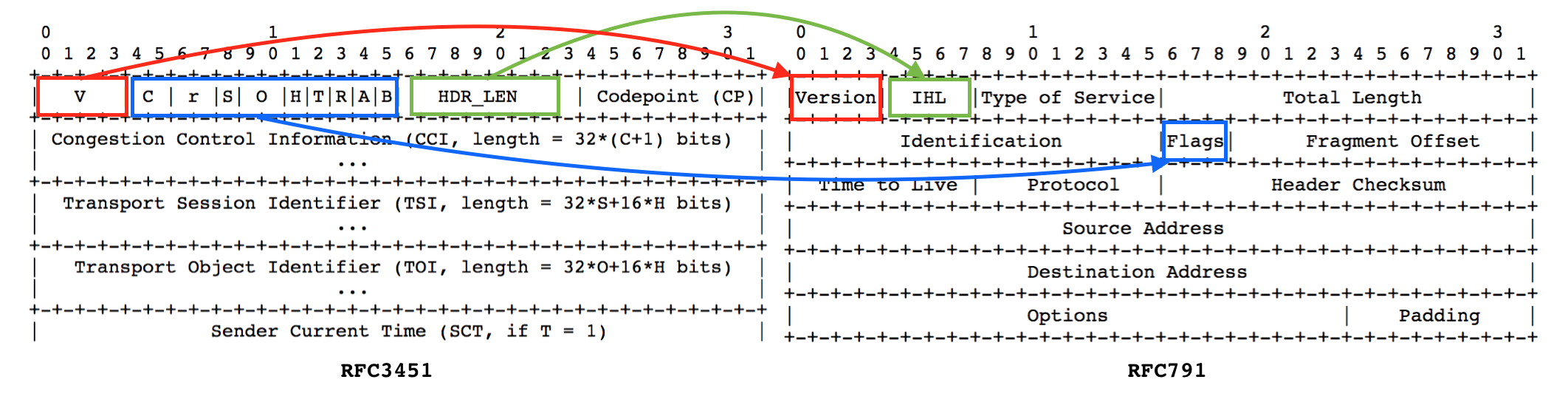}
\caption{Examples of Various Writing Styles in RFCs. {\scriptsize Data frames are extracted from RFC3451 and RFC791. Header field "Verion" is written as Version in RFC791 whereas abbreviated V is used in RFC3451. Header field "Header Length" is written as IHL in RFC791 whereas HDR\_LEN is used in RFC3451. Header field "Flag" is written as Flag in RFC791 whereas every flag bit is displayed in RFC3451.
}}
\label{fig2}
\end{figure}

Internet protocol analysis is an advanced computer networking topic that uses a packet analyzer to capture, view, and understand Internet protocols. These Internet specifications and communications protocols are documented in Request for Comment memorandum (RFCs). RFCs present informative resources for protocol analysis. Entity Linking (EL) recognizes and disambiguates named entities in RFCs and links them to a Protocol Knowledge Base (PKB) and is useful for comprehensive protocol analysis (Fig.~\ref{fig1}). However, RFCs are informational or experimental, and their formats are not standard\cite{ref_lncs7}, as they are written in an informal way. Besides, RFCs are typically released by different institutions or individuals over many years. Hence, various writing styles or standards are also used, making RFC replete with abbreviations, simplifications, and obsolete expressions (Fig.~\ref{fig2}). These characteristics make EL in RFC documents extremely difficult. Since each ontology in the PKB is associated with entities presenting huge discrepancies.
Pre-trained language models\cite{ref_lncs1, ref_lncs2, ref_lncs3} have become a robust way to deal with Entity Linking (EL). They capture rich language information from text by unifying pre-trained language representations and downstream tasks, thus improve accuracy in many NLP applications. Among these models, BERT\cite{ref_lncs1} has been the most prominent one in recent NLP studies. Through the self-attention mechanism, BERT manages to encode bidirectional contextual information on character-level, word-level, and sentence-level, which reduces the discrepancies among single words. Fine-tuning on BERT also demonstrate optimal results in various downstream tasks, including Named Entity Recognition\cite{ref_lncs4}, Text Classification\cite{ref_lncs5} and EL\cite{ref_lncs6}. However, the initial BERT has done its pre-training on generic datasets, such as Wikipedia and Book Corpus, for general-purpose. It does not have preferences towards specific domains. In terms of our research, it lacks protocol-specific knowledge in RFCs. Our experiments have shown that standalone fine-tuned BERT is not adequate to perform highly accurate EL in RFCs.

In this paper, we propose a novel model PEL-BERT to tackle EL in RFCs. The key idea is combining a fine-tuned model with an RFC Domain Model. Experimental results have demonstrated our model achieves an accuracy 72.9\% on our annotated RFC dataset, outperforming all the baselines.
%
Briefly speaking, our contributions include:
\begin{itemize}
\item We design a Protocol Knowledge Base (PKB) as the guideline for protocol Entity Linking.
\item We propose PEL-BERT, which joints a fine-tuned language model with an RFC domain model. It achieves the highest result in EL on our annotated dataset, outperforming all the baselines.
\item We give a comprehensive analysis on the performance of pre-trained language models on descriptive texts and abstract concepts.
\end{itemize}

In the upcoming sections, we first present an overview of related researches, followed by a detailed description of the experiment process and elaboration of evaluation methods. In the end, we give some concluding remarks.
\section{Related Works}
\subsection{Entity Linking}
Entity linking is the task to link entity mentions in text with their corresponding ontologies. Most of EL aims to link mentions to a comprehensive Knowledge Base (KB)\cite{ref_lncs13}. Recent approaches have used neural networks\cite{ref_lncs14, ref_lncs15} to capture the correspondence between a mention’s context and a proposed entity in the KB. Graph-based\cite{ref_lncs16, ref_lncs17} and various joint methods\cite{ref_lncs18, ref_lncs19} are also widely used. Instead of linking to KB, there are also approaches to perform EL to ad-hoc entity lists\cite{ref_lncs20}. In terms of our experiment, we focus on the former one, namely, EL on KB. We aim to bridge the gap between mentions in real word text and entities in well established theoretical schemas. A similar study that links mentions from RFC documents to a list of ontologies is conducted by Jero\cite{ref_lncs24} to generate grammar-based fuzzing. Given limited training data, they generalized this problem by assigning the property with the maximum key phrase overlap to a header field.
\subsection{Fine-tuning BERT for Classification}
BERT demonstrates high accuracy in classification tasks and has been widely applied to many domain-specific fields. Lee\cite{ref_lncs21} proposed a way of fine-tuning BERT for patent classification, outperforming DeepPatent\cite{ref_lncs22}. Adhikari\cite{ref_lncs23} applied BERT to four accessible datasets to perform document classification, improving the baselines in classification. However, there haven't been many studies in classification in protocols using BERT.
\subsection{Learning with Scarce Annotations}
In this paper, we also consider the problem of data-scarcity. Since we manually annotate our dataset, resolving data scarcity allows us to train our model on a relatively small dataset, which largely reduces human efforts. Transfer learning (TL) had been applied widely in researches\cite{ref_lncs26, ref_lncs27}, which uses classifiers trained in large datasets similar to but not the same as target datasets to perform new tasks. Active Learning (AL) is another way of dealing with data scarcity\cite{ref_lncs28, ref_lncs29}. It selects queries or sub-spans that are most informative and reinforces its learning result by iterating. Bootstrapping, where classifiers use their own predictions to teach themselves, is widely used for Entity Set Expansion (ESE)\cite{ref_lncs30, ref_lncs31}. It provides enriching datasets by acquiring new samples each iteration. We utilize AL and Bootstrapping in our experiment.
 \section{Approach}
\begin{figure}[htb]
\setlength{\abovecaptionskip}{0pt}
\setlength{\belowcaptionskip}{10pt}
\includegraphics[width=\textwidth]{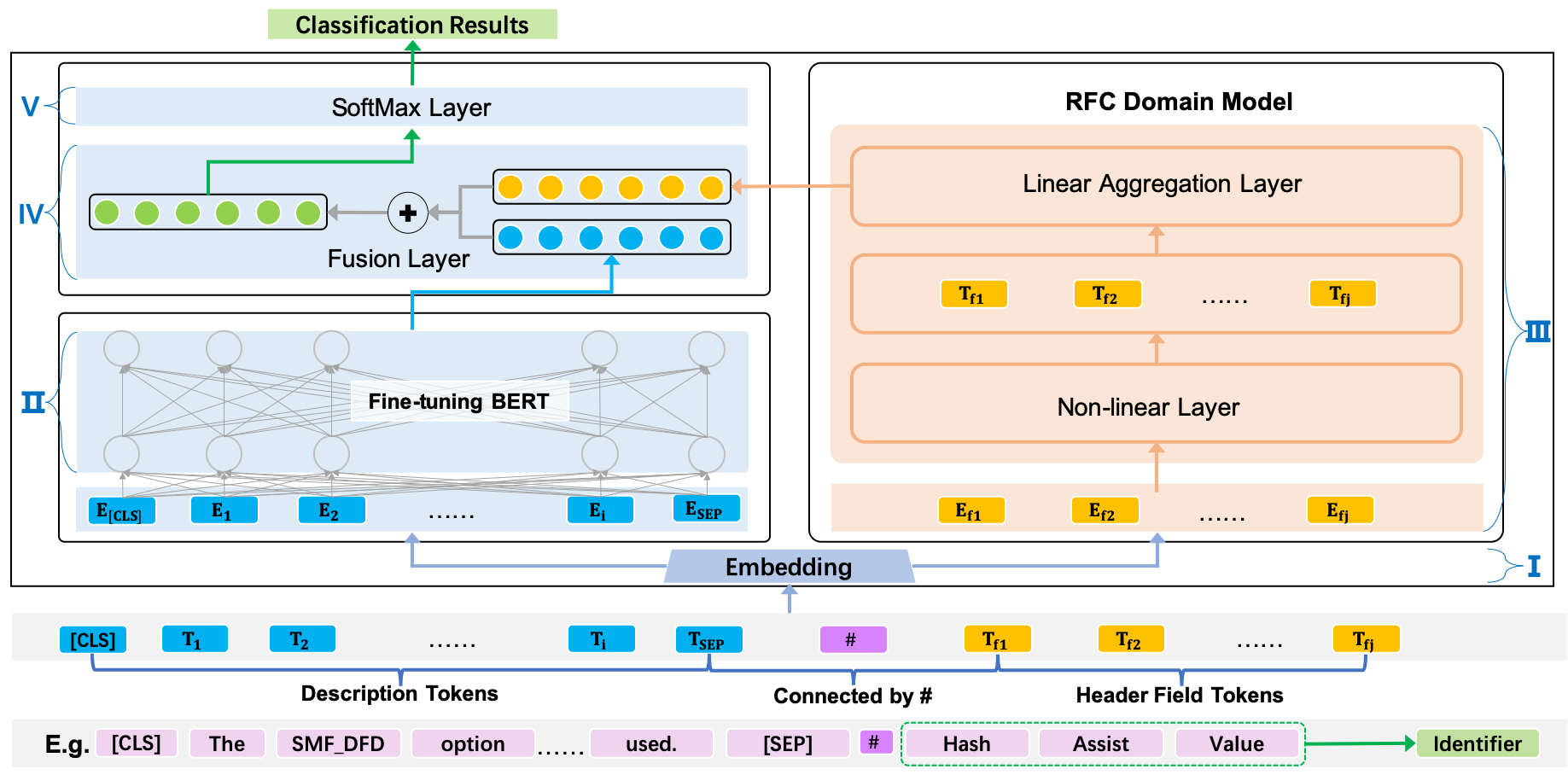}
\caption{PEL-BERT Model Architecture} \label{fig3}
\end{figure}

In this section, we elaborate on the implementation details of our PEL-BERT model. PEL-BERT consists of four modules: Embedding, Fine-tuned BERT Model, RFC Domain Model, and Fusion. The overall architecture of the model is shown in Fig.~\ref{fig3}. In the following subsections, we describe each of these four modules in detail.

{\bfseries \uppercase\expandafter{\romannumeral1}. Embedding:}
The input for each experiment is header descriptions concatenated with header fields. Header Field is parsed through graphs of headers (Fig.~\ref{fig1}) that are highly unanimous across RFCs. Description is the text chunk that references to its corresponding header field. We infer Description form the nearby contexts of its header field using Zero-Shot Learning (ZSL)\cite{ref_lncs25} similar to Jero\cite{ref_lncs24}.

We apply the word list and embedding mechanism proposed in BERT to convert descriptions and header fields into word embeddings. Special tokens [CLS] and [SEP] are concatenated to each description, indicating the start and end of every input sequence, respectively. While header fields are fed directly into the model, descriptions, and header fields are tokenized and then input into the embedding layer where tokens are converted into word embeddings. In the embedding layer, we use the embedding mechanism proposed in BERT. Therefore, each word embedding is constructed by joining the corresponding token, segment, and position embeddings. Since the output of the embedding layer is delivered to two different models, the position embeddings for header fields and descriptions are not consecutive; instead, they all start from zero.

{\bfseries \uppercase\expandafter{\romannumeral2}. Fine-tuned BERT:}
We use BERT$_{BASE}$ as part of our PEL-BERT. The word embeddings for descriptions, denoted as {E$_{(CLS)}$, E$_{1}$, E$_{2}$, ..., E$_{n}$, E$_{(SEP)}$} are fed into BERT. The output, denoted as {T$_{(CLS)}$, T$_{(1)}$, T$_{(2)}$, ..., T$_{(n)}$, T$_{(SEP)}$}, represents the contextual information for descriptions. Let {\itshape info$_{words}$}, {\itshape info$_{code}$} and {\itshape HS$_{E}$} be the descriptions, their embeddings and their hidden states, respectively, this process can be formalized into following formulas:
\begin{equation}
  info_{code} = encoder(info_{words})
\end{equation}
\begin{equation}
  HS_{E} = BERT(info_{code})
\end{equation}
The hidden state, denoted as T$_{CLS}$, of the special token [CLS] is considered as the aggregated sequence representation in the output of the BERT layer, which then sends to the Adding Operation Unit. The following formula is used to compute T$_{CLS}$:
\begin{equation}
  T_{(CLS)} = HS_{E}[0]
\end{equation}

{\bfseries \uppercase\expandafter{\romannumeral3}. RFC Domain Model:}
The word embeddings for header fields denoted as {E$_{F_1}$, E$_{F_2}$, ..., E$_{F_n}$}, are fed into a non-linear layer. We consider BPNN, CNN, Bi-GRU as possible non-linear layers. This non-linear layer converts word embeddings of header fields into hidden states used as the input for the Linear Aggregation Layer. Since BERT is a pre-training model based on large datasets of general knowledge, the semantic information of header fields is minimal compared to the large number of words presented in it. This would reduce the effect of header fields because other trivial words would overshadow them. Through the non-linear layer, we assure that the semantic information of header fields is examined separately, thus preserve the valuable information of header fields. The output for the non-linear layer, denoted as {T$_{F_1}$, T$_{F_2}$ , ..., T$_{F_n}$}, represents header fields. Let {\itshape field$_{words}$}, {\itshape field$_{code}$} and {\itshape HS$_{E}$} be the header fields, their embeddings, and their hidden states, respectively, this process can be formalized into following formulas:
\begin{equation}
  field_{code} = encoder(field_{words})
\end{equation}
\begin{equation}
  HS_{F} = NonLinearModel(field_{code})
\end{equation}
To further incorporate semantic information for header fields, we design a Linear Aggregation Layer to concatenate all the hidden states, the intermediate result from the non-linear layer, to fully explore the augmentation of heuristics inferred from header fields. The Linear Aggregation Layer provides the input for the following Adding Operation Unit. Let T$_{A}$ be the final representation for header fields. This process can be formalized into the following formulas:
\begin{equation}
  T_{A} = linearAGGR(HS_{F})
\end{equation}

{\bfseries \uppercase\expandafter{\romannumeral4}. Fusion Layer:}
The fusion phrase consists of an Adding Operation Unit and a {\itshape softmax} layer. The Adding Operation Unit transforms T$_{A}$ using an activation function \(\sigma\).
This transformation constructs a new representation for header fields that enables element-wise concatenation with T$_{CLS}$. In this unit, we also leverage T$_{A}$ against T$_{CLS}$, so that the result is still dominated by BERT, but also integrates the heuristics of header fields. Through this, the header fields are involved in the fine-tuning process of BERT. We regard this as heuristics between two different models. In our specific experiment settings, {\itshape ReLU} is used as the activation function. During backpropagation, the parameters in BERT and the non-linear layer are mutually independent, which enable BERT to preserve most of its innate characteristics and maintain its high performance. The Adding Operation Unit is applied to combine T$_{CLS}$ and T$_{A}$ to produce a vector representation O$_{A}$ that is finally ready for the classification task. This process can be formalized into the following formulas:
\begin{equation}
O_{A} = \sigma(W \times T_{A} + b) + T_{CLS}
\end{equation}

{\bfseries \uppercase\expandafter{\romannumeral5}. Classification:}
The output of the fusion layer O$_{A}$ is processed by a {\itshape softmax} layer, acquiring semantic labels, denoted as {\itshape pred}, for current inputs. We use Average Cross Entropy as our loss function. This process can be formalized into the following formulas:
\begin{equation}
pred = e^{O_{A}}/{\sum_{j}e^{O_{j}}}
\end{equation}
\begin{equation}
L(O_{A}) = -log\frac{e^{O_{A}}}{\sum_{j}e^{O_{j}}} = -O_{A} + log{\sum_{j}e^{O_{j}}}
\end{equation}
\section{Experiment}
\subsection{Experimental Setup}
{\bfseries Dataset:}
In our experiment, each training sample consists of a header field, a description that describes this header field, and the Knowledge Base entity this header field belongs to, denoted as a triple: {\itshape\{Header Field, Description, Knowledge Base Entity\}} (Fig.~\ref{fig1}). We infer Header Fields and Description from RFCs and manually craft a set of 12 Knowledge Base Entities based on prior knowledge on computer network and protocols.

We sample 71 RFC documents that contain header formats descriptions in their catalogs. We collect a total number of 507 samples and split them into training and test set. After eliminating types that contain too few samples, we mainly consider the following feature set: label, length, content, boolean, address, enumeration set, version number, reserved field, and checksum. The distribution of samples of each feature is shown in Table~\ref{tab1}.
\begin{table}
\setlength{\abovecaptionskip}{-10pt}
\setlength{\belowcaptionskip}{0pt}
\caption{Summary of the Dataset}\label{tab1}
\begin{center}
{\begin{tabular}[l]{@{}lcccccc}
\toprule[0.75pt]
  Category & Size\\
\midrule[0.5pt]
  Identifier-label & 112\\
  Length & 70\\
  Data & 64 \\
  Boolean & 62\\
  Identifier-address & 53 \\
  Enum & 45 \\
  Version Number & 36 \\
  Reserved & 34 \\
  Checksum & 31 \\
\bottomrule[0.75pt]
\end{tabular}}
\end{center}
\label{symbols}
\end{table}

{\bfseries Training:}
All of our models (PEL-BERT-a, PEL-BERT-b, PEL-BERT-c) have 12 transformer blocks, 768 hidden units, and 12 self-attention heads. For PEL-BERT, we first initialize it using BERT$_{BASE}$, then fine-tune the model for six epochs with the learning rate of 2e$^{-5}$. During training and testing, the maximum text length is set to 10 tokens\cite{ref_lncs33}. This limit is chosen because header fields often consist of short phrases in RFCs.
\subsection{Baseline}
In this subsection, we introduce six baselines against which we evaluate our method. We compare the result from PEL-BERT with these six baselines. We tune the hyperparameters for baselines for fair comparison, and all the baselines take the word embedding for header descriptions as their input.

{\itshape SVM\cite{ref_lncs34}:} We use stochastic gradient descent to train the SVM model. The constraints are adjusted using L2 Regularization and margin is set to 1.0. Other parameters are initialized with randomly assigned values.

{\itshape BPNN\cite{ref_lncs35}:} All parameters are initialized with randomly assigned values. The dropout rate is set to 0.1 to avoid over-fitting. During training, we adopt adaptive gradient descent strategy.

{\itshape CNN\cite{ref_lncs36}:} We use 3 kernels during convolution. Kernel size is set to 3 * 768. The size of kernels in the max-pooling phrase is 2 * 2. Batch size is 1. Dropout rate is set to 0.1. The output is sent into a linear layer and a {\itshape softmax} layer to make predictions.

{\itshape Bi-GRU\cite{ref_lncs37}:} Bi-GRU acts similarly to the memory cell in the LSTM network. All parameters are initialized with randomly assigned values. We concatenate the output and use the same linear and {\itshape softmax} lays as CNN to post-process the output.

{\itshape Adhikari et al.\cite{ref_lncs32}:} The model takes word embedding as its input. We concatenate the outputs into a linear layer and a Softmax. The size of kernels in the max-pooling phrase is 2 * 2. The dropout rate is set to 0.1.

{\itshape DocBERT\cite{ref_lncs23}:} A sentence classifier based on BERT. The output of BERT is feed into a linear layer and a {\itshape softmax} layer.

To validate the impact header fields have on the experiment, all the baselines do not consider header information. Also, for SVM, BPNN, CNN, Bi-GRU, we use 8000 iterations to approximate the number of single epoch in BERT. Besides, we set the learning rate for all the baselines except DocBERT to 2e$^{-2}$ for faster convergence.

\section{Evaluation and Analysis}
We report Acc, Avg$_{P}$, Avg$_{R}$, Avg$_{F}$ for all categories in the schema using 10-fold cross-validation. Acc indicates the accuracy upon the training set. Avg$_{P}$, Avg$_{R}$, Avg$_{F}$ indicate the average precision, recall, and F-measure across all the categories in PKB, respectively. For category {\itshape a}, if our model makes the right decision, we do \(TP(a) = TP(a) + 1\), otherwise do \(FP(a) = FP(a) + 1\)and \(FN(b) = FN(b) + 1\). Let {\itshape C} be the total number of categories, {\itshape N} be the total number of samples. We compute the above four criteria using the following formulas:
\begin{gather}
  Acc = \sum_{a=1}^C\frac{TP(a)}{N} \\
  Avg_{P} = \sum_{a=1}^C\frac{TP(a)}{TP(a) + FP(a)} \\
  Avg_{R} = \sum_{a=1}^C\frac{TP(a)}{TP(a) + FN(a)} \\
  Avg_{F} = \frac{2 \times Avg_{P} \times Avg_{R}}{Avg_{P} + Avg_{R}}
\end{gather}
%
\subsection{Evaluation Between Different Implementations of PEL-BERT}
Regarding possible implementations of PEL-BERT, we evaluate PEL-BERT based on three non-linear layers, namely, BPNN, CNN, and Bi-GRU. The statistics are shown in Table~\ref{tab2}.
\begin{table}[htbp]
\setlength{\abovecaptionskip}{-10pt}
\setlength{\belowcaptionskip}{0pt}
\caption{Detailed results of Acc, Avg$_{P}$, Avg$_{R}$, Avg$_{F}$ are shown. Best results are highlighted in bold font. Training is done on our manully annotated RFC dataset.}\label{tab2}
\begin{center}
\begin{tabular}{lcccccccccc}
\specialrule{0.1em}{3pt}{3pt}
\tabincell{c}{Exp. Group} & & & Model & & & Acc & Avg$_{P}$ & Avg$_{R}$ & Avg$_{F}$
& \tabincell{c}{Learning Rate}\\
\specialrule{0.07em}{1.5pt}{1.5pt}
\multirow{3}{*}{\tabincell{c}{\bfseries Our \\ \bfseries Approach}}
& & & \tabincell{c}{PEL-BERT-a (BPNN)} & & & 72.4\% & {\bfseries 73.9\%} & 74.3\% & 74.1\% & 2e$^{-5}$\\
& & & \tabincell{c}{PEL-BERT-b (CNN)} & & & 49.6\% & 51.3\% & 53.0\% & 52.1\% & 2e$^{-5}$\\
& & & \tabincell{c}{PEL-BERT-c (Bi-GRU)} & & & {\bfseries 72.9\%} & 73.7\% &
{\bfseries 74.7\%} & {\bfseries 74.2\%} & 2e$^{-5}$\\
\specialrule{0.1em}{1pt}{1pt}
\end{tabular}
\end{center}
\end{table}
As the result shows, PEL-BERT based on Bi-GRU achieves the best result on our dataset, reaching the highest accuracy of 72.9\% and highest Avg$_{F}$ of 74.2\%. Since CNN is insufficient to capture contextual information, its result is inferior to BPNN and Bi-GRU, which are better in utilizing the location embeddings in the word vector. Bi-GRU is derived from RNN, which is suitable to deal with segmented information. Therefore, we choose Bi-GRU as the non-linear layer for RFC Domain Model in PEL-BERT.
\subsection{Ablation Study: Evaluation on performance of the Joint Model}
This set of experiments is designed to evaluate the impact header fields have on classification, namely, whether the RFC Domain Model contributes to increasing Avg$_{F}$. The statistics are shown in Table~\ref{tab3}.
\begin{table}[htbp]
\setlength{\abovecaptionskip}{-10pt}
\setlength{\belowcaptionskip}{0pt}
\caption{Detailed results of Acc, Avg$_{P}$, Avg$_{R}$, Avg$_{F}$ are shown. Best results are highlighted in bold font. Training is done on our manully annotated RFC dataset.}\label{tab3}
\begin{center}
\begin{tabular}{lcccccccccc}
\specialrule{0.1em}{3pt}{3pt}
\tabincell{c}{Exp. Group} & & & Model & & & Acc & Avg$_{P}$ & Avg$_{R}$ & Avg$_{F}$
& \tabincell{c}{Learning Rate}\\
\specialrule{0.07em}{1.5pt}{1.5pt}
BERT & & & Fine-tuned BERT & & & 69.8\% & 72.7\% & 72.2\% & 72.4\% & 2e$^{-5}$\\
\specialrule{0.05em}{1.5pt}{1.5pt}
\multirow{3}{*}{\tabincell{c}{RFC Domain \\ Model$^{\rm *}$}}
& & & BPNN + linearAGGR & & & 47.3\% & 50.2\% & 48.5\% & 49.4\% & 2e$^{-5}$\\
& & & CNN + linearAGGR& & & 35.5\% & 33.5\% & 34.9\% & 34.2\% & 2e$^{-5}$\\
& & & Bi-GRU + linearAGGR & & & 60.8\% & 64.1\% & 64.1\% & 64.1\% & 2e$^{-5}$\\
\specialrule{0.05em}{1.5pt}{1.5pt}
{\tabincell{c}{\bfseries Joint \\ \bfseries Model}} & & & \tabincell{c}{PEL-BERT-c (Bi-GRU)} & & & {\bfseries 72.9\%} & {\bfseries 73.7\%} &
{\bfseries 74.7\%} & {\bfseries 74.2\%} & 2e$^{-5}$\\
\specialrule{0.1em}{1pt}{1pt}
\end{tabular}
\footnotesize{$^*$ Inputs for RFC Domain Models are word embeddings for header descriptions.}
\end{center}
\end{table}
The accuracy of fine-tuned BERT is 69.8\%. Individual RFC Domain Model also achieves inferior results. Whereas when RFC Domain Mode is aggregated with BERT, the accuracy reaches 72.9\%, which is higher than merely using BERT or RFC Domain Model. We can infer that the fusion of the domain model and fine-tuned language model can improve the performance by injecting domain-specific knowledge. Therefore, we choose PEL-BERT-c (Bi-GRU) as our final model.
\subsection{Comparison with Baselines}
From the statistics shown in Table~\ref{tab4}, our approach, the joint model of BERT and RFC Domain Model, achieves the best accuracy of 72.9\%.
\begin{table}[htbp]
\setlength{\abovecaptionskip}{-10pt}
\setlength{\belowcaptionskip}{-0pt}
\caption{Detailed results of Acc, Avg$_{P}$, Avg$_{R}$, Avg$_{F}$ are shown. Best results are highlighted in bold font. Training is done on our manully annotated RFC dataset.}\label{tab4}
\begin{center}
\begin{tabular}{lcccccccccc}
\specialrule{0.1em}{3pt}{3pt}
\tabincell{c}{Exp. Group} & & & Model & & & Acc & Avg$_{P}$ & Avg$_{R}$ & Avg$_{F}$
& \tabincell{c}{Learning Rate}\\
\specialrule{0.07em}{1.5pt}{1.5pt}
\multirow{6}{*}{Baseline}
& & & SVM & & & 10.8\% & 10.4\% & 10.8\% & 10.6\% & 2e$^{-2}$\\
& & & BPNN & & & 55.8\% & 47.8\% & 48.7\% & 48.2\% & 2e$^{-2}$\\
& & & CNN & & & 48.0\% & 44.5\% & 45.2\% & 44.8\% & 2e$^{-2}$\\
& & & Bi-GRU & & & 53.6\% & 44.9\% & 41.3\% & 43.0\% & 2e$^{-2}$\\
& & & Adhikari\cite{ref_lncs32} & & & 57.6\% & 48.3\% & 48.3\% & 48.3\% & 2e$^{-2}$\\
& & & DocBERT\cite{ref_lncs23} & & & 70.6\% & 71.8\% & 71.1\% & 71.4\% & 2e$^{-5}$\\
\specialrule{0.05em}{1.5pt}{1.5pt}
{\tabincell{c}{\bfseries Our \\ \bfseries Approach}}
& & & PLE-BERT & & & {\bfseries 72.9\%} & {\bfseries 73.7\%} & {\bfseries 74.7\%} & {\bfseries 74.2\%} & 2e$^{-5}$\\
\specialrule{0.1em}{1pt}{1pt}
\end{tabular} \\
\end{center}
\end{table}
\subsection{Analysis}
We can draw an analogy between language models pre-trained on generic corpora and human beings. These models achieve decent results for descriptive texts because they manage to encode contextual information. However, our experimental results illustrate that they fail to comprehend abstract concepts like {\itshape header field}. For example, the header field “IHL” (Fig.~\ref{fig1}) is an abstract concept in the protocol domain. Domain-specific information cannot be inferred from its lexical presentation. Thus we are unable to link it to specific categories KB. In our approach, we not only use domain knowledge to fine-tune BERT but also design a domain model to learn domain-specific knowledge from these header fields explicitly. Finally, we combine these two models as our PEL-BERT model. Compared with baselines, our model achieves the best result.

\section{Discussion}
{\bfseries Use RFC Domain Model to Deal with Header Fields:} Given the fact that BERT is apt for texts with sequential relations. By merely appending header fields to the descriptions, we attached false information which will mislead BERT to regard the header fields and descriptions appearing in the adjacent context, while they are not. This hinders BERT's performance. Therefore, concatenating header fields directly to descriptions is not appropriate. As a result, we do not solely apply BERT to our EL task, with header fields concatenated with descriptions as its input.

{\bfseries Use Neural Network to Handle Header Fields:} We consider using neural network, such as Bi-GRU and BPNN, as the domain model to deal with header fields independently, rather than Pre-training Model because our RFC dataset is minimal compared to the large corpora BERT is pre-trained on. The heuristic knowledge BERT acquired from generic corpora largely compromises the valuable domain-specific information contained in our dataset. By training a model explicitly for header fields, we manage to exploit the domain knowledge contained in them fully.

\section{Conclusion and Future Work}
In this paper, we propose PEL-BERT to better fuse domain-specific knowledge into general purposed language models. We use PEL-BERT to link RFC entities to Protocol Knowledge Base. We give an comprehensive analysis on the performance of pre-trained language models on descriptive texts and abstract concepts. The experimental results demonstrate that PEL-BERT has better abilities in Entity Linking than all the baselines.
There are two points we want to address in future studies: (1) optimize and extend out dataset to cover a border range of RFCs; (2) evaluate our model on other domain-specific datasets other than protocols to prove its universality. Resolving these problems will lead to a more comprehensive language understanding.

%

\end{document}